%% file: main.tex
\documentclass[11pt,a4paper]{article}
\usepackage{emnlp2020}
\usepackage{times}
\usepackage{latexsym}

\usepackage{microtype}

\usepackage[utf8]{inputenc}
\usepackage{bbold}
\usepackage{url,graphicx,tabularx,array, booktabs, mathrsfs, color, epstopdf, tabu, svg, diagbox, mathtools, multirow} 
\usepackage{amsmath, amssymb, bm, cases, enumitem}
\usepackage{algorithm}
\usepackage[noend]{algpseudocode}

\newcommand{\finetune}{\emph{Finetune}}
\newcommand{\full}{\emph{Full}}

\newcommand{\herding}{\emph{ER$_{herding}$}}
\newcommand{\random}{\emph{ER$_{random}$}}
\newcommand{\loss}{\emph{ER$_{prio}$}}

\newcommand{\Ltwo}{\emph{L2}}
\newcommand{\distillation}{\emph{KD}}
\newcommand{\dropout}{\emph{Dropout}}
\newcommand{\arer}{\emph{ARPER}}

\newcommand{\twostar}{$^{\star\star}$}
\newcommand{\onestar}{$^{\star}$}

\aclfinalcopy % Uncomment this line for the final submission
\def\aclpaperid{2525} %  Enter the acl Paper ID here

\title{Continual Learning for Natural Language Generation in Task-oriented Dialog Systems}

\author{
Fei Mi$^1$,
Liangwei Chen$^1$,
Mengjie Zhao$^2$,
Minlie Huang$^3$\and
Boi Faltings$^1$\\
$^1$LIA, EPFL, Lausanne, Switzerland\\
$^2$CIS, LMU, Munich, Germany \\
$^3$CoAI, DCST, Tsinghua University, Beijing, China
\\
\texttt{\{fei.mi,liangwei.chen,boi.faltings\}@epfl.ch}\\
\texttt{mzhao@cis.lmu.de}, \quad
\texttt{aihuang@tsinghua.edu.cn}
}

\begin{document}
\maketitle

\input{text/abs}

\input{text/intro}

\input{text/related}

\input{text/model}

\input{text/exp}

\input{text/conclusion}

\bibliographystyle{acl_natbib}
\bibliography{ref}
\clearpage

\input{text/appendix}

\end{document}

%% file: text/abs.tex
\begin{abstract}

Natural language generation (NLG) is an essential
component of task-oriented dialog systems. Despite the recent success of neural approaches for NLG, they are typically developed in an offline manner for particular domains. 
To better fit real-life applications where new data come in a stream, we study NLG in a ``continual learning'' setting to expand its knowledge to new domains or functionalities incrementally.
The major challenge towards this goal is catastrophic forgetting, meaning that a continually trained model tends to forget the knowledge it has learned before.
To this end, we propose a method called \arer\ (Adaptively Regularized Prioritized Exemplar Replay) by replaying prioritized historical exemplars, together with an adaptive regularization technique based on Elastic Weight Consolidation.
Extensive experiments to continually learn new domains and intents are conducted on MultiWoZ-2.0 to benchmark \arer\ with a wide range of techniques. 
Empirical results demonstrate that \arer\ significantly
outperforms other methods by effectively mitigating the detrimental catastrophic forgetting issue.

\end{abstract}

%% file: text/intro.tex
\section{Introduction}

As an essential part of task-oriented dialog systems \cite{wen2016toward,bordes2016learning}, the task of Natural Language Generation (NLG) is to produce a natural language utterance containing the desired information given a \textit{semantic representation} (so-called dialog act).
Existing NLG models \cite{wen2015semantically,tran2017natural,tseng2018variational} are typically trained offline using annotated data from a single or a fixed set of domains. However,
a desirable dialog system in real-life applications often needs to expand its knowledge to new domains and functionalities.
Therefore, it is crucial to develop an NLG approach with the capability of continual
learning after a dialog system is deployed.
Specifically, an NLG model should be able to continually learn new utterance patterns without forgetting the old ones it has already learned.

The major challenge of continual learning lies in catastrophic forgetting \cite{mccloskey1989catastrophic,french1999catastrophic}. Namely, a neural model trained on new data tends to forget the knowledge it has acquired on previous data.
We diagnose in Section \ref{subsec:diagnosis} that neural NLG models suffer such detrimental catastrophic forgetting issues when continually trained on new domains.
A naive solution is to retrain the NLG model using all historical data every time.
However, it is not scalable due to severe computation and storage overhead.

To this end, we propose storing a small set of representative utterances from previous data, namely \textit{exemplars}, and replay them to the NLG model each time it needs to be trained on new data. 
Methods using exemplars have shown great success in different continual learning \cite{rebuffi2017icarl,castro2018end,chaudhry2019continual} and reinforcement learning \cite{schaul2015prioritized,andrychowicz2017hindsight} tasks.
In this paper, we propose a \textit{prioritized} exemplar selection scheme to choose representative and diverse exemplar utterances for NLG. We empirically demonstrate that the prioritized exemplar replay helps to alleviate catastrophic forgetting by a large degree. 

In practice, the number of exemplars should be reasonably small to maintain a manageable memory footprint. Therefore, the constraint of not forgetting old utterance patterns is not strong enough. 
To enforce a stronger constraint, we propose a regularization method based on the well-known technique, Elastic Weight Consolidation (EWC \cite{kirkpatrick2017overcoming}). The idea is to use a quadratic term to elastically regularize the parameters that are important for previous data.
Besides the wide application in computer vision, EWC has been recently applied to the domain adaptation task for Neural Machine Translation \cite{thompson2019overcoming,saunders2019domain}.
In this paper, we combine EWC with exemplar replay by approximating the Fisher Information Matrix w.r.t. the carefully chosen exemplars so that not all historical data need to be stored. 
Furthermore, we propose to adaptively adjust the regularization weight to consider the difference between new and old data to flexibly deal with different new data distributions.

To summarize our contribution, 
(1) to the best of our knowledge, this is the first attempt to study the practical continual learning configuration for NLG in task-oriented dialog systems;
(2) we propose a method called Adaptively Regularized Prioritized Exemplar Replay (\arer) for this task, and benchmark it with a wide range of state-of-the-art continual learning techniques;
(3) extensive experiments are conducted on the MultiWoZ-2.0 \cite{budzianowski2018multiwoz} dataset to continually learn new tasks, including domains and intents using two base NLG models. Empirical results demonstrate the superior performance of \arer\ and its ability to mitigate catastrophic forgetting. Our code is available at \url{https://github.com/MiFei/Continual-Learning-for-NLG}

%% file: text/related.tex
\section{Related Work}

%%%一般区分 人做主语和论文做主语；citet

 \paragraph{Continual Learning.}
 The major challenge for continual learning is catastrophic forgetting~\cite{mccloskey1989catastrophic,french1999catastrophic}, where optimization over new data leads to performance degradation on data learned before. Methods designed to mitigate catastrophic forgetting fall into three categories: \emph{regularization}, \emph{exemplar replay}, and \emph{dynamic architectures}. 
Methods using dynamic architectures~\cite{rusu2016progressive,maltoni2019continuous} increase model parameters throughout the continual learning process, which leads to an unfair comparison with other methods. 
In this work, we focus on the first two categories.

Regularization methods add specific regularization terms to consolidate knowledge learned before. \citet{li2017learning} introduced the knowledge distillation~\cite{hinton2015distilling} to penalize model logit change, and it has been widely employed in~\citet{rebuffi2017icarl,castro2018end,wu2019large,hou2019learning,zhao2019maintaining}. Another direction is to regularize parameters crucial to old knowledge according to various importance measures~\cite{kirkpatrick2017overcoming,zenke2017continual,aljundi2018memory}.

Exemplar replay methods store past samples, a.k.a \textit{exemplars}, and replay them periodically.
Instead of selecting exemplars at random, \citet{rebuffi2017icarl} incorporated the \textit{Herding} technique~\cite{welling2009herding} to choose exemplars that best approximate the mean feature vector of a class, and it is widely used in~\citet{castro2018end,wu2019large,hou2019learning,zhao2019maintaining,mi2020generalized,mi2020ader}. 
\citet{ramalho2019adaptive} proposed to store samples that the model is least confident.
\citet{chaudhry2019continual} demonstrated the effectiveness of exemplars for various continual learning tasks in computer vision.

\paragraph{Catastrophic Forgetting in NLP.} The catastrophic forgetting issue in NLP tasks has raised increasing attention recently \cite{mou2016transferable,chronopoulou2019embarrassingly}. 
\citet{yogatama2019learning,arora2019does} identified the detrimental catastrophic forgetting issue while fine-tuning ELMo \cite{peters2018deep} and BERT \cite{devlin2019bert}.
To deal with this issue, \citet{he2019mix} proposed to replay pre-train data during fine-tuning heavily, and \citet{chen2020recall} proposed an improved Adam optimizer to recall knowledge captured during pre-training.
The catastrophic forgetting issue is also noticed in domain adaptation setups for neural machine translation \cite{saunders2019domain,thompson2019overcoming,varis2019unsupervised} and the reading comprehension task \cite{xu2019forget}. 
%\mf{\cite{saunders2020reducing} later applied EWC \cite{kirkpatrick2017overcoming} to retain the gender balance of translated sentences during fine-tuning.}

\citet{lee2017toward} firstly studied the continual learning setting for dialog state tracking in task-oriented dialog systems. However, their setting is still a one-time adaptation process, and the adopted dataset is small.
\citet{shen2019progressive} recently applied progressive network \cite{rusu2016progressive} for the semantic slot filling task from a continual learning perspective similar to ours. However, their method is based on a dynamic architecture that is beyond the scope of this paper.
\citet{liu2019continual} proposed a Boolean operation of ``conceptor'' matrices for continually learning sentence representations using linear encoders.
\citet{li2020compositional} combined continual learning and language systematic compositionality for sequence-to-sequence learning tasks.

\paragraph{Natural Language Generation (NLG).} In this paper, we focus on NLG for task-oriented dialog systems.
A series of neural methods have been proposed to generate accurate, natural,
and diverse utterances, including HLSTM~\cite{wen2015stochastic}, SCLSTM~\cite{wen2015semantically}, Enc-Dec~\cite{wen2016toward}, RALSTM~\cite{tran2017natural}, SCVAE~\cite{tseng2018variational}. 

Recent works have considered the domain adaptation setting.
	\citet{tseng2018variational,tran2018dual} proposed to learn domain-invariant representations using VAE~\cite{kingma2013auto}. 
	They later designed two domain adaptation critics with an adversarial training algorithm~\cite{tran2018adversarial}.
Recently, \citet{mi2019meta,qian2019domain,peng2020few} studied learning new domains with limited training data.
However, existing methods only consider a one-time adaptation process. The continual learning setting and the corresponding catastrophic forgetting issue remain to be explored.

%% file: text/model.tex
\section{Model}

In this section, we first introduce the background of neural NLG models in Section \ref{subsec:background}, and the continual learning formulation in Section \ref{subsec:formulation}. In Section \ref{subsec:arer}, we introduce the proposed method \arer.

\subsection{Background on Neural NLG Models}
\label{subsec:background}

The NLG component of task-oriented dialog systems is to produce natural language utterances conditioned on a \textit{semantic representation} called dialog act~(DA). Specifically, the dialog act $\mathbf{d}$ is defined as the combination of \textit{intent} $\mathbf{I}$ and a set of slot-value pairs $S(\mathbf{d}) = \{(s_i, v_i)\}_{i=1}^{p}$:
\begin{equation}
\mathbf{d} = [\underbrace{ \mathbf{I} }_\text{Intent}, \underbrace{(s_1, v_1), \dots, (s_p, v_p)}_\text{Slot-value pairs} ],
\label{eq:da}
\end{equation}
where $p$ is the number of slot-value pairs.  Intent $\mathbf{I}$ controls the utterance functionality, while slot-value pairs contain information to express.  For example, ``\textit{There is a restaurant called [La Margherita] that serves [Italian] food.}'' is an utterance corresponding to a DA ``\textit{[Inform, (name=La Margherita, food=Italian)]}''

	Neural models have recently shown promising results for NLG tasks.
	Conditioned on a DA, a neural NLG model generates an utterance containing the desired information word by word.
    For a DA $\mathbf{d}$ with the corresponding ground truth utterance $\mathbf{Y} = (y_1, y_1, ..., y_{K})$, the probability of generating $\mathbf{Y}$ is factorized as below:
	\vspace{-0.05in}
	\begin{equation}
	f_{\theta}(\mathbf{Y}, \mathbf{d}) = \prod_{k=1}^{K} p_{y_k} =\prod_{k=1}^{K} p(y_{k} | y_{<k}, \mathbf{d}, \theta),  %\vspace{-0.02in}
	\end{equation}

where $f_{\theta}$ is the NLG model parameterized by $\theta$, and $p_{y_k}$ is the output probability (i.e. softmax of logits) of the ground truth token $y_k$ at position $k$.
   The typical objective function for an utterance $\mathbf{Y}$ with DA $\mathbf{d}$ is the average cross-entropy loss w.r.t. all tokens in the utterance \cite{wen2015semantically,wen2016toward,tran2017natural,peng2020few}: 
    \begin{equation}
    L_{CE}(\mathbf{Y}, \mathbf{d} , \theta)  = - \frac{1}{K} \sum_{k=1}^{K} log(p_{y_k})
    \end{equation}

%%% divided by K is “length normalization”
\begin{figure}[t!]
		\centering
		\includegraphics[width=0.498\textwidth]{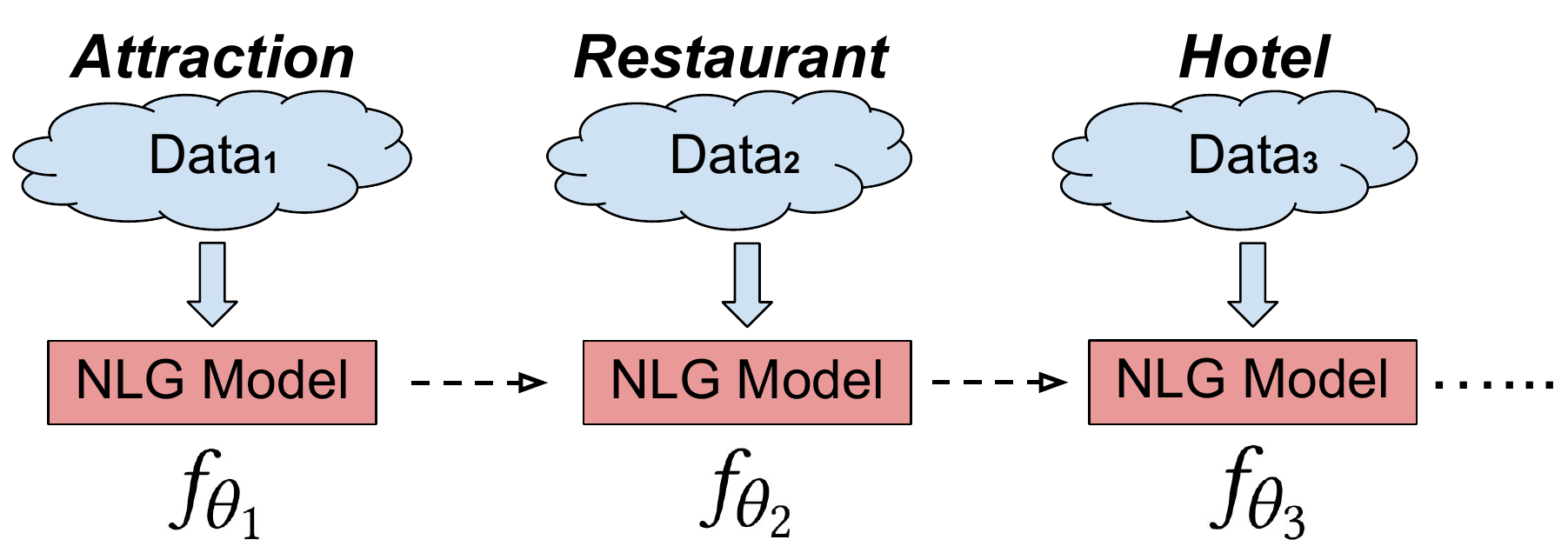} 
 \vspace{-0.13in} \caption{An example for a NLG model to continually learn new domains.  The model needs to perform well on all domains it has seen before. For example $f_{\theta_3}$ needs to deal with all three previous domains (\textit{Attraction, Restaurant, Hotel}).} \vspace{-0.13in}
        \label{fig:setting}
\end{figure}

\vspace{-0.05in}
\subsection{Continual Learning of NLG}
\label{subsec:formulation}
    In practice, an NLG model needs to continually learn new domains or functionalities.
    Without loss of generality, we assume that new data arrive \textit{task by task} \cite{rebuffi2017icarl,kirkpatrick2017overcoming}. In a new task $t$, new data $\mathbf{D}_t$ are used to train the NLG model $f_{\theta_{t-1}}$ obtained till the last task. The updated model $f_{\theta_t}$ needs to perform well on \textit{all} tasks so far.
    An example setting of continually learning new domains is illustrated in Figure \ref{fig:setting}.
    A task can be defined with different modalities to reflect diverse real-life applications. 
	In subsequent experiments, we consider continually learning new domains and intents in Eq. (\ref{eq:da}).

We emphasize that the setting of continual learning is different from that of domain adaptation. The latter is a one-time adaptation process, and the focus is to optimize performance on a target domain transferred from source domains but without considering potential performance drop on source domains \cite{mi2019meta,qian2019domain,peng2020few}. In contrast, continual learning requires a NLG model to continually learn new tasks in multiple transfers, and the goal is to make the model perform well on all tasks learned so far.

\subsection{Adaptively Regularized Prioritized Exemplar Replay (\arer)}
\label{subsec:arer}

We introduce the proposed method (\arer) with prioritized exemplar replay and an adaptive regularization technique to further alleviate the catastrophic forgetting issue.

\subsubsection{Prioritized Exemplar Replay}

To prevent the NLG model catastrophically forgetting utterance patterns in earlier tasks, a small subset of a task's utterances are selected as \textit{exemplars}, and exemplars in previous tasks are \textit{replayed} to the later tasks. 
During training the NLG model $f_{\theta_t}$ for task $t$, the set of exemplars in previous tasks, denoted as $\mathbf{E}_{1:t-1}=\{ \mathbf{E}_{1}, \dots, \mathbf{E}_{t-1} \}$, is replayed by joining with the data $\mathbf{D}_t$ of the current task. 
Therefore, the training objective with exemplar replay can be written as:
\begin{equation}
L_{ER}(\theta_t) =   \sum_{\{\mathbf{Y}, \mathbf{d} \} \in \mathbf{D}_t \cup \mathbf{E}_{1:t-1} } L_{CE}(\mathbf{Y}, \mathbf{d} , \theta_t) .
\label{eq:er}
\end{equation}
\vspace{-0.1in}

The set of exemplars of task $t$, referred to as $\mathbf{E}_t$, is selected after $f_{\theta_t}$ has been trained and will be replayed to later tasks.

The quality of exemplars is crucial to preserve the performance on previous tasks. We propose a \textit{prioritized} exemplar selection method to select representative and diverse utterances as follows.

\paragraph{Representative utterances.} 
The first criterion is that exemplars $\mathbf{E}_t$ of a task $t$ should be representative of $\mathbf{D}_t$.
We propose to select $\mathbf{E}_t$ as a \textit{priority list} from $\mathbf{D}_t$ that \textit{minimize} a priority score:
\begin{equation}
U(\mathbf{Y}, \mathbf{d}) =  \ L_{CE}(\mathbf{Y}, \mathbf{d} , \theta_t) \cdot|S(\mathbf{d})|^\beta,
\label{eq:utility}
\end{equation}
where $S(\mathbf{d})$ is the set of slots in $\mathbf{Y}$, and $\beta$ is a hyper-parameter.
This formula correlates the representativeness of an utterance to its $L_{CE}$.
Intuitively, the NLG model $f_{\theta_t}$ trained on $\mathbf{D}_t$ should be confident with representative utterances of $\mathbf{D}_t$, i.e., low $L_{CE}$.
However, $L_{CE}$ is agnostic to the number of slots.
We found that an utterance with many common slots in a task could also have very low $L_{CE}$, yet using such utterances as exemplars may lead to overfitting and thus forgetting of previous general knowledge.
The second term $|S(\mathbf{d})|^\beta$ controls the importance of the number of slots in an utterance to be prioritized as exemplars. 
We empirically found in Appendix \ref{appendix:model} that the best $\beta$ is greater than 0. 

\begin{algorithm}[!t]
\caption{\textit{ARPER.select\_exemplars}: Prioritized exemplar selection procedure of \arer\ for task \textit{t}}\label{alg:euclid}
\begin{algorithmic}[1]
\Procedure{\textit{select\_exemplars}}{$\mathbf{D}_t, f_{\theta_t}, m$}
\State $\mathbf{E}_t \gets$ \textbf{new} $Priority\_list()$
\State $\mathbf{D}_t \gets$ $sort( \mathbf{D}_t, key=\textit{U}, order=asc)$
\While{$|\mathbf{E}_t| < m$}
\State $\mathbf{S}_{seen} \gets \textbf{new}$ $Set()$
\For{$\{ \textbf{Y}, \textbf{d} \}  \in \textbf{D}_t$}
\If{$S(\mathbf{d}) \in \mathbf{S}_{seen}$} continue
\Else{} 
\State $\mathbf{D}_t.remove( \{\mathbf{Y}, \mathbf{d} \})$ 
\State $\mathbf{E}_t.insert( \{\mathbf{Y}, \mathbf{d} \})$
\State $\mathbf{S}_{seen}.insert( S(\mathbf{d}))$ 
\If { $|\mathbf{E}_t| == m$} 
\State \textbf{return} $\mathbf{E}_t$
\EndIf
\EndIf
\EndFor
\EndWhile \label{euclidendwhile}
\EndProcedure
\end{algorithmic}
\end{algorithm}

\paragraph{Diverse utterances.} 
The second criterion is that exemplars should contain diverse slots of the task, rather than being similar or repetitive.
A drawback of the above priority score is that similar or duplicated utterances containing the same set of frequent slots could be prioritized over utterances w.r.t. a diverse set of slots.
To encourage diversity of selected exemplars, we propose an iterative approach to add data from $\mathbf{D}_t$ to the priority list $\mathbf{E}_t$ based on the above priority score. At each iteration, if the set of slots of the current utterance is already covered by utterances in $\mathbf{E}_t$, we skip it and move on to the data with the next best priority score.

Algorithm 1 shows the procedure to select $m$ exemplars as a priority list $\mathbf{E}_t$ from $\mathbf{D}_t$. The outer loop allows multiple passes through $\mathbf{D}_t$ to select various utterances for the same set of slots $S(\mathbf{d})$.

\subsubsection{Reducing Exemplars in Previous Tasks}
Algorithm 1 requires the number of exemplars to be given. A straightforward choice is to store the same and fixed number of exemplars for each task as in \citet{castro2018end,wu2019large,hou2019learning}. However, there are two drawbacks in this method: (1). the memory usage increases linearly with the number of tasks; (2) it does not discriminate tasks with different difficulty levels.

To this end, we propose to store a \textit{fixed} number of exemplars throughout the entire continual learning process to maintain a bounded memory footprint as in \citet{rebuffi2017icarl}. 
As more tasks are continually learned, exemplars in previous tasks are gradually reduced by only keeping the ones in the \textit{front} of the priority list\footnote{the priority list implementation allows reducing exemplars in constant time for each task}, and the exemplar size of a task is set to be proportional to the training data size of the task to differentiate the task's difficulty.
To be specific, suppose $M$ exemplars are kept in total. The number of exemplars for a task is:
\begin{equation}
|\mathbf{E}_i| = M \cdot \frac{|\mathbf{D}_i|}{ \sum_{j=1}^t |\mathbf{D}_j| }, \forall i \in 1, \dots, t,
\label{eq:reduce}
\end{equation}
where we choose 250/500 for $M$ in experiments.

\subsubsection{ Constraint with Adaptive Elastic Weight Consolidation}

Although exemplars of previous tasks are stored and replayed, the size of exemplars should be reasonably small ($M \ll |\mathbf{D}_{1:t}|$) to reduce memory overhead. As a consequence, the constraint we have made to prevent the NLG model catastrophically forgetting previous utterance patterns is not strong enough.
To enforce a stronger constraint, we propose a regularization method based on the well-known Elastic Weight Consolidation (EWC, \citealp{kirkpatrick2017overcoming}) technique.

\paragraph{Elastic Weight Consolidation (EWC).}

EWC utilizes a quadratic term to elastically regularize parameters important for previous tasks.
The loss function of using the EWC regularization together with exemplar replay for task $t$ can be written as:
\vspace{-0.06in}
\begin{equation}
L_{ER\_EWC}(\theta_t) = L_{ER}(\theta_t) + \lambda \sum_i^N F_i (\theta_{t,i} - \theta_{t-1,i})^2
\label{eq:ewc}
\end{equation}
where $N$ is the number of model parameters;  $\theta_{t-1,i}$ is the $i$-th converged parameter of the model trained till the previous task; $F_i = \nabla^2 L_{CE}^{\mathbf{E}_{1:t-1}}(\theta_{t-1,i})$ is the $i$-th diagonal element of the Fisher Information Matrix approximated w.r.t. the set of previous exemplars $\mathbf{E}_{1:t-1}$. $F_i$ measures the importance of $\theta_{t-1,i}$ to previous tasks represented by $\mathbf{E}_{1:t-1}$.
Typical usages of EWC compute $F_i$ w.r.t. a uniformly sampled subset from historical data. Yet, we propose to compute $F_i$ w.r.t. the carefully chosen $\mathbf{E}_{1:t-1}$ so that not all historical data need to be stored.
The scalar $\lambda$ controls the contribution of the quadratic regularization term.
The idea is to elastically penalize changes on parameters important (with large $F_i$) to previous tasks, and more plasticity is assigned to parameters with small $F_i$.

\paragraph{Adaptive regularization.}
In practice, new tasks have different difficulties and similarities compared to previous tasks. Therefore, the degree of need to preserve the previous knowledge varies.
To this end, we propose an adaptive weight ($\lambda$) for the EWC regularization term as follows:
\begin{equation}
\lambda = \lambda_{base} \sqrt{V_{1:t-1} / V_t},
\label{eq:lambda}
\end{equation}
where $V_{1:t-1}$ is the \textit{old} word vocabulary size in previous tasks, and $V_t$ is the \textit{new} word vocabulary size in the current task $t$; $\lambda_{base}$ is a hyper-parameter. In
general, $\lambda$ increases when the ratio of the size of old
word vocabularies to that of new ones increases. In other words, the regularization term becomes more important when the new task contains fewer new vocabularies to learn.

\begin{algorithm}[!t]
\caption{\textit{ARPER.learn\_task}: Procedure of \arer\ to continually learn task \textit{t}}
%\caption{\textit{APRER}: Continually learn task \textit{t}}
\begin{algorithmic}[1]
\Procedure{\textit{learn\_task}}{$\mathbf{D}_t, \mathbf{E}_{1:t-1}, f_{\theta_{t-1}}, M$}
\State $\theta_t \gets \theta_{t-1}$
\While{$\theta_t$ not converged}
\State $\theta_t \gets update(L_{ER\_EWC}(\theta_t))$
\EndWhile
\State $m \gets M \cdot \frac{|\mathbf{D}_t|}{\Sigma_{j = 1}^{t}|\mathbf{D}_j|}$
\State $\mathbf{E}_t \gets \textit{ARPER.select\_exemplars}(\mathbf{D}_t, f_{\theta_t}, m)$
\For {$j = 1\; \mathbf{to } \; t - 1$}
\State $\mathbf{E}_j \gets \mathbf{E}_j.top(M \cdot  \frac{|\mathbf{D}_j|}{\Sigma_{j = 1}^{t}|\mathbf{D}_j|})$
\EndFor
\State \textbf{return } $\textit{f}_{\theta_t}, \mathbf{E}_t$
\EndProcedure
\end{algorithmic}
\end{algorithm}

Algorithm 2 summarizes the continual learning procedure of \arer\ for task $t$. $\theta_t$ is initialized with $\theta_{t-1}$, and it is trained with prioritized exemplar replay and adaptive EWC in Eq. (\ref{eq:ewc}). After training $\theta_t$, exemplars $\mathbf{E}_t$ of task $t$ are computed by Algorithm 1, and exemplars in previous tasks are reduced by keeping the most prioritized ones to preserve the total exemplar size.

%% file: text/exp.tex
\section{Experiments}

\subsection{Dataset}
\label{subsec:dataset}
We use the MultiWoZ-2.0 dataset \footnote{extracted for NLG at \url{https://github.com/andy194673/nlg-sclstm-multiwoz}} \cite{budzianowski2018multiwoz} containing six domains (\textit{Attraction, Hotel, Restaurant, Booking, Taxi and Train}) and seven DA intents (\textit{``Inform, Request, Select, Recommend, Book, Offer-Booked, No-Offer''}).
The original train/validation/test splits are used. For methods using exemplars, both training and validation set are continually expanded with exemplars extracted from previous tasks. 

\begin{figure}[t!]
		\centering
		\includegraphics[width=0.45\textwidth]{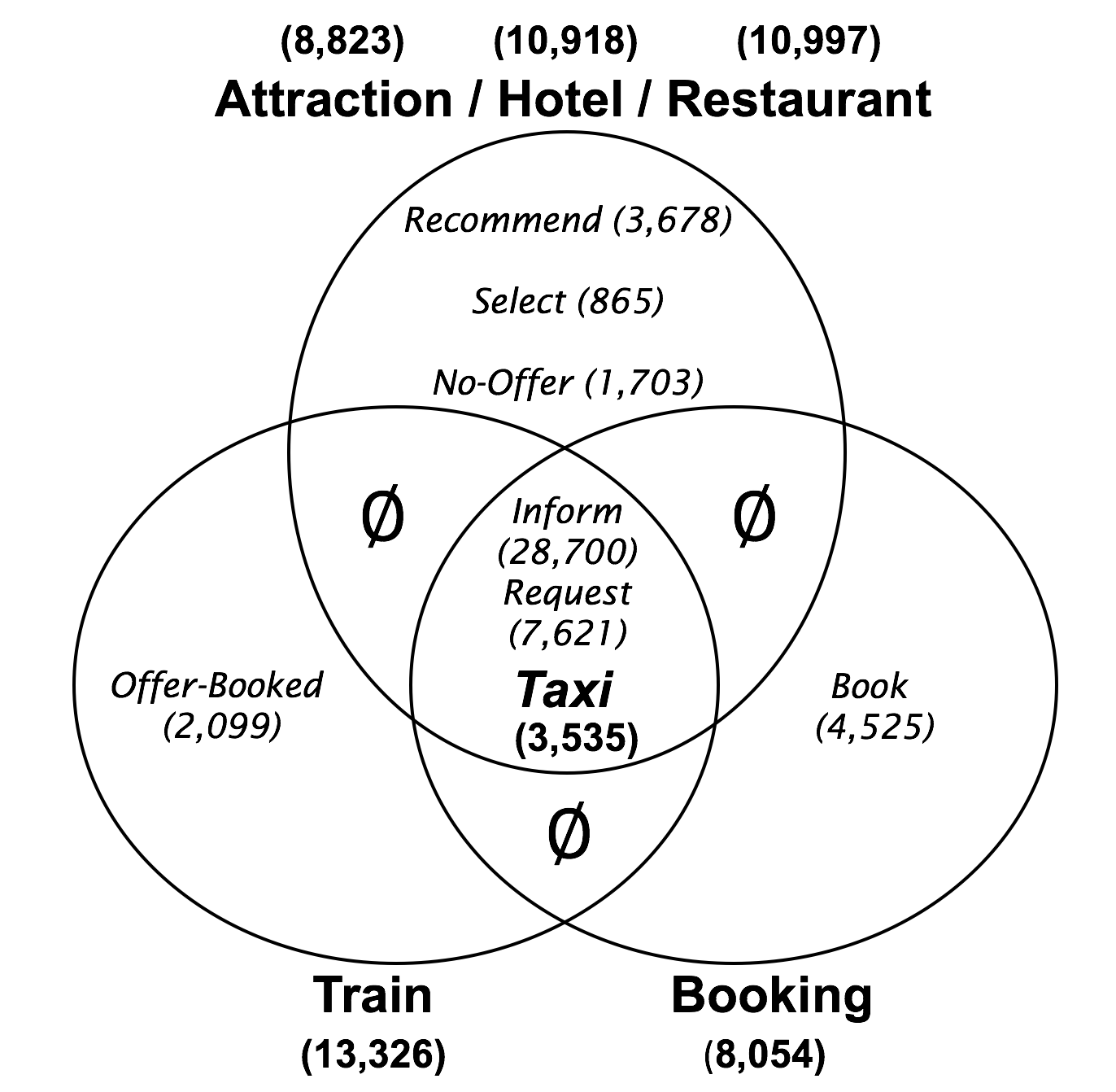} 
       \vspace{-0.05in} \caption{ Venn diagram visualizing intents in different domains. The number of utterances of each domain (bold) and intents (italic) is indicated in parentheses.} \vspace{-0.05in}
        \label{fig:stat}
\end{figure}

To support experiments on continual learning new domains, we pre-processed the original dataset by segmenting multi-domain utterances into single-domain ones.
For instance, an utterance \textit{``The ADC Theatre is located on Park Street. Before I find your train, could you tell me where you would like to go?''} is split into two utterances with domain ``\textit{Attraction}'' and ``\textit{Train}'' separately. 
If multiple sentences of the same domain in the original utterance exist, they are still kept in one utterance after pre-processing.
In each continual learning task, all training data of one domain are used to train the NLG model, as illustrated in Figure \ref{fig:setting}.
Similar pre-processing is done at the granularity of DA intents for experiments in Section \ref{subsec:da}.
The statistics of the pre-processed MultiWoZ-2.0 dataset is illustrated in Figure \ref{fig:stat}.
The resulting datasets and the pre-processing scripts are open-sourced. 

\subsection{Evaluation Metrics}

Following previous studies, we use the slot error rate (SER) and the BLEU-4
score \cite{papineni2002bleu} as evaluation metrics. SER is the ratio of the number of \textit{missing} and \textit{redundant} slots in a generated utterance to the total number of ground truth slots in the DA.

To better evaluate the continual learning ability, we use two additional commonly used metrics \cite{kemker2018measuring} for both SER and BLEU-4:
\begin{equation*}
\Omega_{all} = \frac{1}{T} \sum_{i=1}^T \Omega_{all,i}, \quad
\Omega_{first} = \frac{1}{T} \sum_{i=1}^{T} \Omega_{first,i} 
%\Omega_{last} & = \frac{1}{T-1} \sum_{i=1}^T \Omega_{last,i}
\end{equation*}
where T is the total number of continual learning tasks; 
$\Omega_{all,i}$ is the test performance on \emph{all} the tasks after the $i^{th}$ task has been learned;
$\Omega_{first,i}$ is that on the \emph{first} task after the $i^{th}$ task has been learned.
Since $\Omega$ can be either SER or BLEU-4, both $\Omega_{all}$ and $\Omega_{first}$ have two versions.
$\Omega_{all}$ evaluates the overall performance, while $\Omega_{first}$ evaluates the ability to alleviate catastrophic forgetting.

\subsection{Baselines}

Two methods without exemplars are as below: 

\begin{itemize}[itemsep=0pt,topsep=2pt,leftmargin=12pt]
\item \textbf{\finetune}: At each task, the NLG model is initialized with the model obtained till the last task, and then fine-tuned with the data from the current task. 

\item \textbf{\full}: At each task, the NLG model is trained with the data from the current and \textit{all} historical tasks. This is the ``upper bound'' performance for continual learning w.r.t. $\Omega_{all}$.
\end{itemize}

Several exemplar replay (\textit{ER}) methods trained with Eq. (\ref{eq:er}) using different exemplar selection schemes are compared:

\begin{itemize}[itemsep=0pt,topsep=2pt,leftmargin=12pt]
    \item \textbf{\herding}~\cite{welling2009herding,rebuffi2017icarl}: This scheme chooses exemplars that best approximate the mean DA vector over all training examples of this task. 
\item \textbf{\random}: This scheme selects exemplars at random. Despite its simplicity, the distribution of the selected exemplars is the same as the distribution of the current task in expectation.
\item \textbf{\loss}: The proposed prioritized scheme (c.f. Algorithm 1) to select representative and diverse exemplars.
\end{itemize}

Based on \loss, four regularization methods (including ours) to further alleviate catastrophic forgetting are compared:

\input{table/domain_fixed_new}
\begin{itemize}[itemsep=0pt,topsep=2pt,leftmargin=12pt]
\item\textbf{ \Ltwo}: A static L2 regularization by setting $F_i = 1$ in Eq. (\ref{eq:ewc}). It regularizes all parameters equally. %It corresponds to the weight decay used by \cite{saunders2019domain,he2019mix}
\item \textbf{\distillation}~\cite{rebuffi2017icarl,wu2019large,hou2019learning}: The widely-used knowledge distillation (KD) loss \cite{hinton2015distilling} is adopted by distilling the prediction logit of current model w.r.t. the prediction logit of the model trained till the last task. More implementation details are included in Appendix \ref{appendix:model}.
\item \textbf{\dropout}~\cite{mirzadeh2020dropout}: Dropout \citet{hinton2012improving} is recently shown by \cite{mirzadeh2020dropout} that it effectively alleviates catastrophic forgetting. We tuned different dropout rates assigned to the non-recurrent connections. 

\item \textbf{\arer}\ (c.f. Algorithm 2): The proposed method using adaptive EWC with \loss.
\end{itemize}

We utilized the well-recognized semantically-conditioned
LSTM (SCLSTM \citealp{wen2015semantically}) as the base NLG model $f_{\theta}$ \footnote{Comparisons based on other base NLG models are included in Section \ref{subsec:scave}.} with one hidden layer of size 128.
Dropout is not used by default, which is evaluated as a separate regularization technique (c.f. \loss+\dropout).
For all the above methods, the learning rate of Adam is set to 5e-3, batch size is set to 128, and the maximum number of epochs used to train each task is set to 100.
Early stop to avoid over-fitting is adopted when the validation loss does not decrease for 10 consecutive epochs.
To fairly compare different methods, they are trained with the identical configuration on the first task to have a consistent starting point.
%Experiments have also been conducted using another base NLG model, namely SCVAE \cite{tseng2018variational}, and results are illustrated in Appendix \ref{appendix:scave} due to page limit.
Hyper-parameters of different methods are included in Appendix \ref{appendix:model}.

\begin{figure}[t!]
		\centering
		\includegraphics[width=0.2375\textwidth]{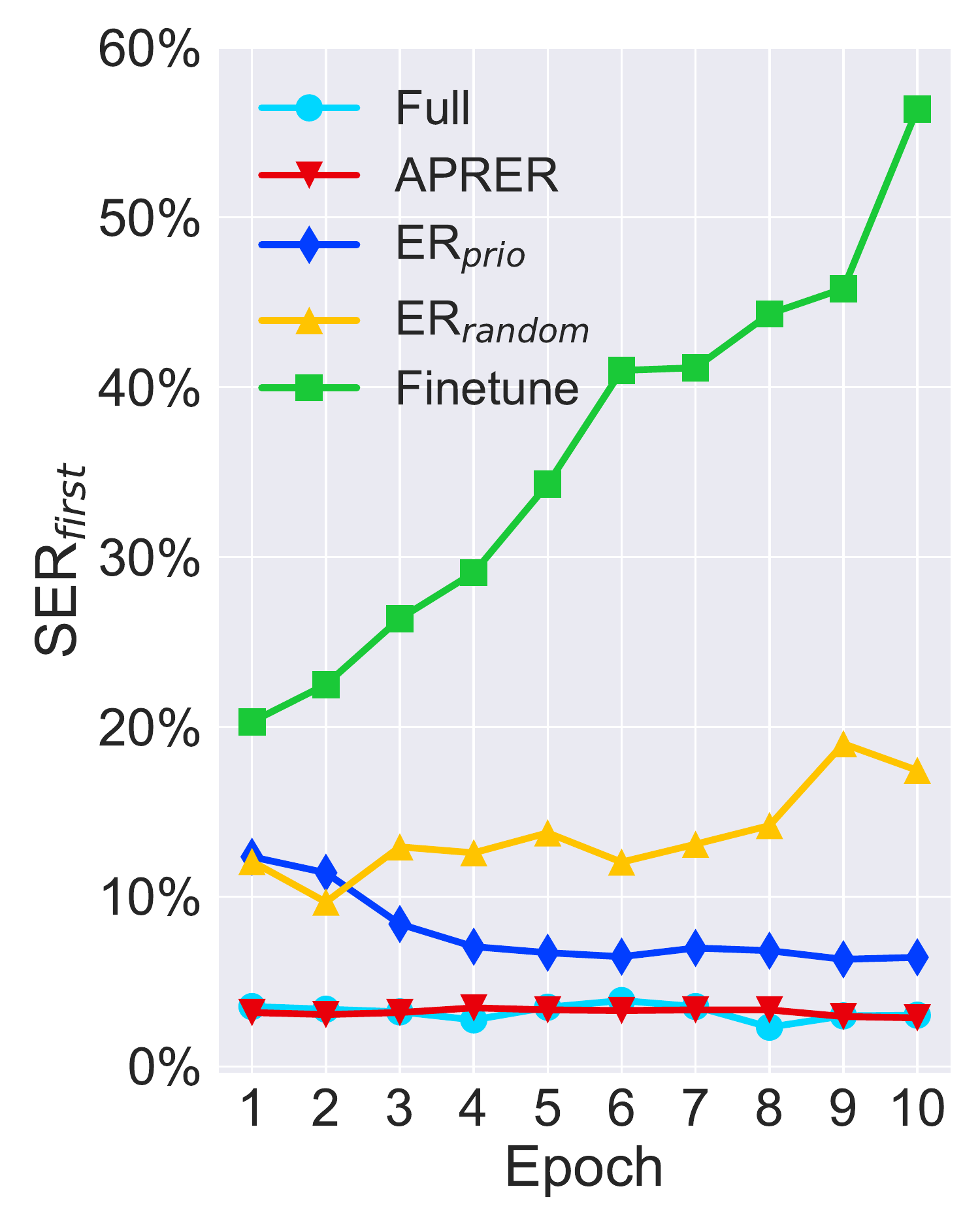}
        \includegraphics[width=0.2375\textwidth]{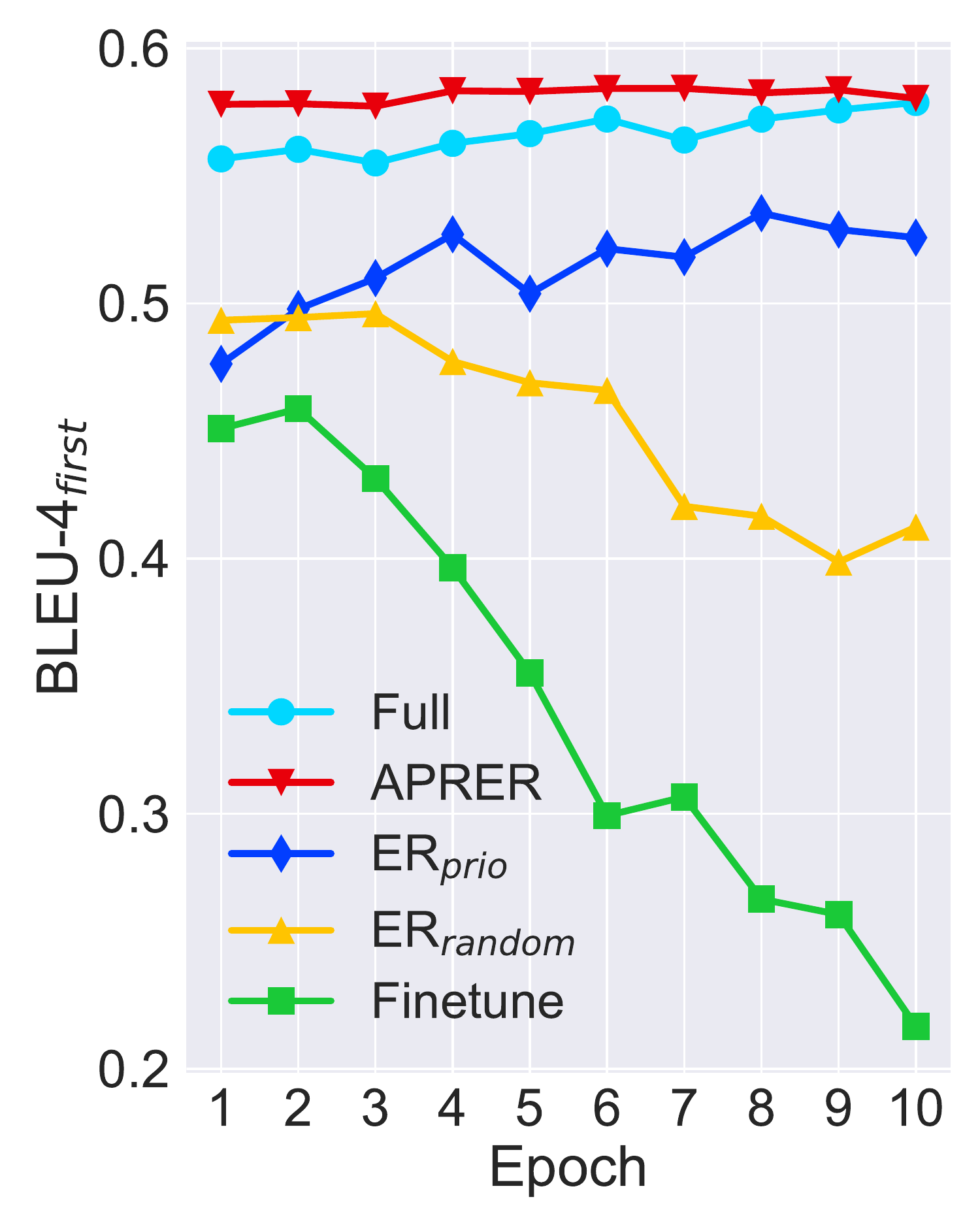}
  \vspace{-0.14in}\caption{Diagnose the catastrophic forgetting issue in NLG. SER (\textbf{Left}) and BLEU-4 (\textbf{Right}) on the test data of ``\textit{Attraction}'' at different epochs when a model pre-trained on the ``\textit{Attraction}'' domain  is continually trained on another ``\textit{Train}'' domain. } \vspace{-0.12in}
        \label{fig:diagnosis}
\end{figure}

\subsection{Diagnose Forgetting in NLG}
\label{subsec:diagnosis}
Before proceeding to our main results, we first diagnose whether the catastrophic forgetting issue exists when training an NLG model continually.
As an example, a model pre-trained on the ``\textit{Attraction}'' domain is continually trained on the ``\textit{Train}'' domain. We present test performance on ``\textit{Attraction}'' at different epochs in Figure \ref{fig:diagnosis} with 250 exemplars.

We can observe: (1) catastrophic forgetting indeed exists as indicated by the sharp performance drop of \finetune; (2) replaying carefully chosen exemplars helps to alleviate catastrophic forgetting by a large degree, and \loss\ does a better job than \random; and (3) \arer\ greatly mitigates catastrophic forgetting by achieving similar or even better performance compared to \full.

\subsection{Continual Learning New Domains}
\label{subsec:domain}
In this experiment, the data from six domains are presented sequentially.
We test 6 runs with different domain order permutations. Each domain is selected as the first task for one time, and the remaining five domains are randomly ordered \footnote{Exact domain orders are provided in Appendix \ref{appendix:order}}.
Results averaged over 6 runs using 250 and 500 total exemplars are presented in Table \ref{table:domain_overall}.
Several interesting observations can be noted:

% \begin{figure*}[hbt!]
% 		\centering
% 		\includegraphics[width=0.44\textwidth]{figures/domain_average_cul_se.pdf} \qquad
%         \includegraphics[width=0.44\textwidth]{figures/domain_average_se.pdf} 
%  \vspace{-0.1in}  \caption{SER on all seen domains (\textbf{left}) and on the first domain (\textbf{right}) as the number of task increases when continually learning 6 domains using 250 exemplars. Results in each task are averaged over 6 runs.}  
%  \vspace{-0.05in}
%         \label{fig:domain}
% \end{figure*}

\begin{itemize} [itemsep=-1pt,topsep=1pt,leftmargin=12pt]

\item %%%%Performing well on all seen tasks is more difficult than only on the first task. 
All methods except \finetune\ perform worse on all seen tasks ($\Omega_{all}$) than on the first task ($\Omega_{first}$). This is due to the diverse knowledge among different tasks, which increases the difficulty of handling all the tasks. \finetune\ performs poorly in both metrics because of the detrimental catastrophic forgetting issue.

\item Replaying exemplars helps to alleviate the catastrophic forgetting issue. Three \emph{ER} methods substantially outperform \finetune. Moreover, the proposed prioritized exemplar selection scheme is effective, indicated by the superior performance of \loss\ over \herding\ and \random.

 \item \arer\ significantly outperforms three \emph{ER} methods and other regularization-based baselines. Compared to the three closest competitors,  \arer\ is significantly better with $p$-value $< 0.05$ w.r.t SER.

\item The improvement margin of \arer\ is significant w.r.t SER that is critical for measuring an output's fidelity to a given dialog act.
Different methods demonstrate similar performance w.r.t BLEU-4, where several of them approach \full, thus are very close to the upper bound performance.

\item \arer\ achieves
comparable performance w.r.t to the upper bound (\full) on all seen tasks ($\Omega_{all}$) even with a very limited number of exemplars. Moreover, it outperforms \full\ on the first task ($\Omega_{first}$), indicating that \arer\ better mitigates forgetting the first task than \full, and the latter is still interfered by data in later domains.

\end{itemize}

\begin{figure}[t!]
		\centering
		\includegraphics[width=0.47\textwidth]{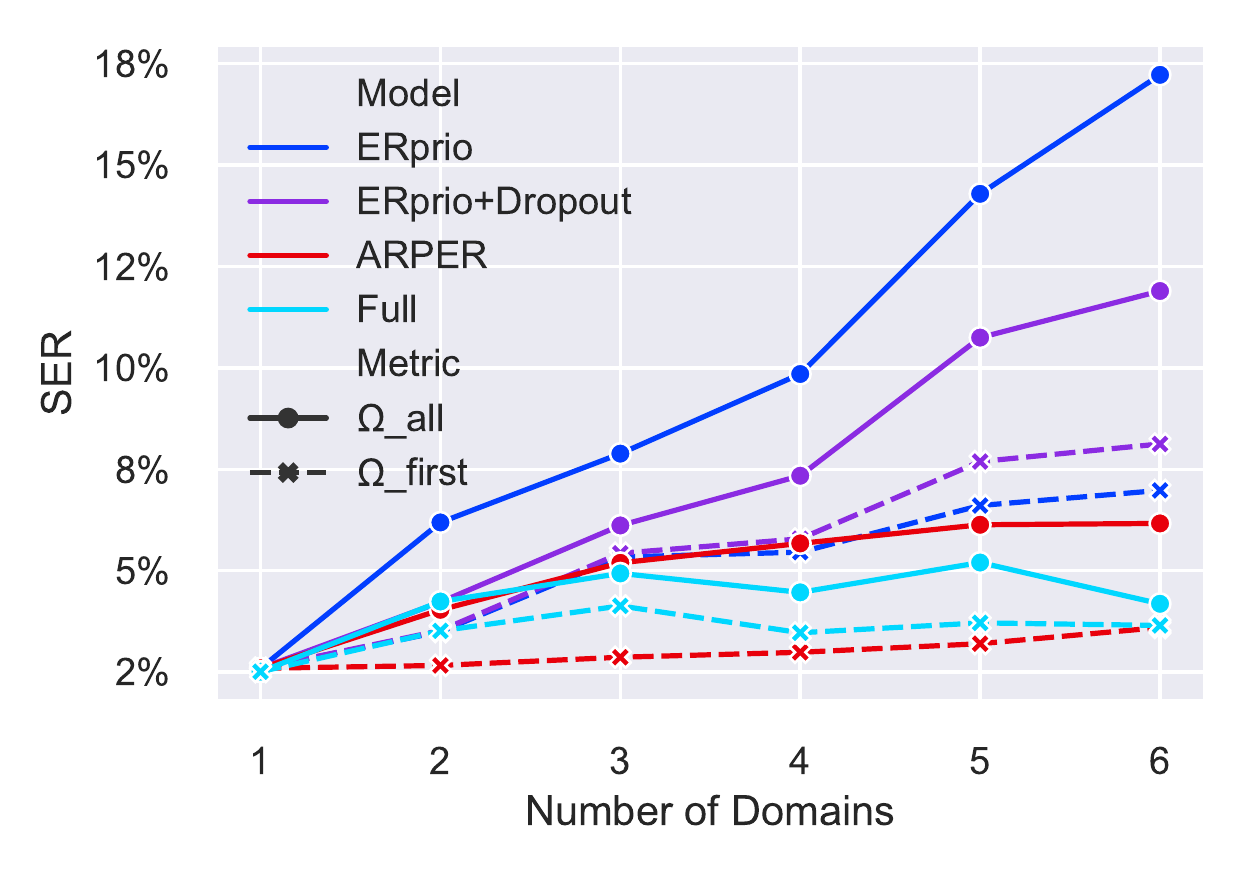} 
 \vspace{-0.15in}  \caption{SER on all seen domains (\textbf{solid}) and on the first domain (\textbf{dashed}) when more domains are continually learned using 250 exemplars.}  \vspace{-0.14in}
        \label{fig:domain}
\end{figure}

\paragraph{Dynamic Results in Continual Learning}
In Figure \ref{fig:domain}, several representative methods are compared as more domains are continually learned.
With more tasks continually learned, \arer\ performs consistently better than other methods on all seen tasks (solid lines), and it is comparable to \full.
On the first task (dashed lines), \arer\ outperforms all the methods, including \full, at every continual learning step. 
These results illustrate the advantage of \arer\ through the entire continual learning process.

\input{table/da}

%We can observe that \arer\ is consistently better than other methods at each individual task. Furthermore, the performance of \arer\ does not degrade much as more tasks are continually learned. In fact, \arer's performance is comparable to that of \full\ on all seen tasks (solid lines), and it better than \full\ on the first task (dashed lines).

\subsection{Continual Learning New DA Intents}
\label{subsec:da}
It is also essential for a task-oriented dialog system to continually learn new functionalities, namely, supporting new DA intents. 
To test this ability, the data of seven DA intents are presented sequentially in the order of decreasing data size, i.e., \textit{``Inform, Request, Book, Recommend, Offer-Booked, No-Offer, Select''}.
Results using 250 exemplars are presented in Table \ref{table:da}. We can observe that \arer\ still largely outperforms other methods, and similar observations for \arer\ can be made as before.
Therefore, we conclude that \arer\ is able to learn new functionalities continually.

Compared to previous experiments, the performance of \loss+\distillation\ degrades, while the performance of \loss+\Ltwo\ improves due to the very large data size in the first task (``\textit{Inform}''), which means that they are sensitive to task orders.

\input{table/ablation}

\input{table/generation_examples}

\subsection{Ablation Study}
In Table \ref{table:ablation}, we compare several simplified versions of \arer\ to understand the effects of different components. Comparisons are based on continually learning 6 domains staring with ``\textit{Attraction}''. 
 We can observe that:  (1). $L_{ER}$ is beneficial because dropping it (``w/o ER'') degrades the performance of \arer. (2). Using prioritized exemplars is advantageous because using random exemplars (``w/o PE'') for \arer\ impairs its performance. (3). Adaptive regularization is also effective, indicated by the superior performance of \arer\ compared to using fixed regularization weights (``w/o AR'').

\subsection{Case Study}

Table \ref{table:samples} shows two examples generated by \arer\ and the closest competitor (\loss+\dropout) on the first domain (“\textit{Attraction}”) after the NLG model is continually trained on all 6 domains starting with ``\textit{Attraction}''.
In both examples, \loss+\dropout\ fails to generate slot ``Fee'' or ``Type'', instead, it mistakenly generates slots belonging to later domains (``\textit{Hotel}'' or ``\textit{Restaurant}'') with several obvious redundant repetitions colored in purple. It means that the NLG model is \textit{interfered} by utterance patterns in later domains, and it forgets some old patterns it has learned before.
In contrast, \arer\ succeeds in both cases without forgetting previously learned patterns.

\subsection{Results using Other NLG Models}
\label{subsec:scave}

\input{table/scvae}

% In this experiment, we changed the base NLG model from SCLSTM to SCVAE \cite{tseng2018variational}.
% Hyper-parameters of SCVAE are set by default\footnote{\url{https://github.com/andy194673/nlg-scvae}}, except that the learning rate is set to 2e-3. 
% Results of using 250 exemplars to continually learn 6 domains starting with ``\textit{Attraction}'' are presented in Table \ref{table:scvae}.
% In general, the relative performance patterns of different methods are similar to that we observed in Section \ref{subsec:domain} and \ref{subsec:da}. 
% Therefore, we can claim that the superior performance of \arer\ can generalize to different base NLG models.

In this experiment, we changed the base NLG model From SCLSTM to SCVAE \cite{tseng2018variational} and GPT-2 \cite{radford2019language}.
For GPT-2, we used the pre-trained model with 12 layers and 117M parameters. As in \citet{peng2020few}, exact slot values are not replaced by special placeholders during training as in SCLSTM and SCVAE. The dialog act is concatenated with the corresponding utterance before feeding into GPT-2. More details are included in Appendix \ref{appendix:model}.

Results of using 250 exemplars to continually learn 6 domains starting with ``\textit{Attraction}'' are presented in Table \ref{table:scvae}. Thanks to the large-scale pre-trained language model, GPT-2 suffers less from the catastrophic forgetting issue because of the better performance of \finetune.
In general, the relative performance patterns of different methods are similar to that we observed in Section \ref{subsec:domain} and \ref{subsec:da}. 
Therefore, we can claim that the superior performance of \arer\ can generalize to different base NLG models.

%% file: table/domain_fixed_new.tex
    \begin{table*}[htb!]
    \fontsize{10}{12}\selectfont
		\centering
		\begin{tabular}{l cc cc cc cc }
			\hline
            &  \multicolumn{4}{c}{\textbf{250 exemplars in total}}  & \multicolumn{4}{c}{\textbf{500 exemplars in total}} \\
            & \multicolumn{2}{c}{$\Omega_{all}$}  & \multicolumn{2}{c}{$\Omega_{first}$} & \multicolumn{2}{c}{$\Omega_{all}$}  & \multicolumn{2}{c}{$\Omega_{first}$} \\
            \cmidrule(r){2-3} \cmidrule(r){4-5} \cmidrule(r){6-7} \cmidrule(r){8-9}
            & SER\% & BLEU-4 & SER\% & BLEU-4 & SER\% & BLEU-4  & SER\% & BLEU-4  \\
              \hline 
              \finetune & 64.46 &	0.361 &	107.27 &	0.253 	 & 64.46 &	0.361  &	107.27  &	0.253  \\
            \herding  & 16.89 &	0.535 &	9.89 &	0.532  & 12.25 &	0.555 &	4.53 &	0.568  \\
            \random  & 10.93 &	0.552 &	6.96 &	0.553 & 8.36 & 0.569 &	4.41 &	0.572 \\
           \loss  & {\quad9.67\twostar} &	0.578 &	{\quad5.28\twostar} &	0.578  & {\quad7.48\twostar} &	0.597 &	{   \hspace{0.012cm} 3.59\onestar} &	0.620  \\
        
           \loss+\Ltwo & 14.94 & 0.579 &	{\quad5.31\twostar} &	0.587  & 10.51 &	0.596 &	{\quad4.28\twostar} &	0.605 \\
           \loss+\distillation & {\quad8.65\twostar} &	0.586 &	6.87 &	0.601	& {\quad7.37\twostar} &	0.596 &	4.89 &	0.617 \\
            \loss+\dropout & {\quad7.15\twostar} &	0.588 &	{\quad5.53\twostar} &	0.594 & {\hspace{0.12cm} 6.09\onestar} &	0.595 &	{\quad4.51\twostar} &	0.616 \\
            \arer & \textbf{ 5.22} &	\textbf{0.590} &	\textbf{ 2.99} &	\textbf{0.624}  & \textbf{      5.12} &	\textbf{0.598} &	\textbf{ 2.81} &	\textbf{0.627} \\
            \hline
            \full & { 4.26} &	0.599 &	{ 3.60} &	0.616  & {    4.26} &	0.599 &	{ 3.60} &	0.616\\
			\hline
		\end{tabular}
		\vspace{-0.05in} \caption{Average performance of continually learning 6 domains using 250/500 exemplars. Best Performance excluding ``\full'' are in bold in each column. In each column , \onestar\ indicates $p<0.05$ and \twostar\ indicates $p<0.01$ for a one-tailed t-test comparing \arer\ to the three top-performing competitors except \full.} \vspace{-0.05in}
        \label{table:domain_overall}
	\end{table*}

%% file: table/da.tex
 \begin{table}[t!]
 \fontsize{10}{12}\selectfont
 \setlength{\tabcolsep}{3pt}
		\centering
		\begin{tabular}{l cc cc }
			\hline
           % &  \multicolumn{4}{c}{\textbf{250 exemplars in total}}  \\
            & \multicolumn{2}{c}{$\Omega_{all}$}  & \multicolumn{2}{c}{$\Omega_{first}$}  \\
            \cmidrule(r){2-3} \cmidrule(r){4-5} 
            & SER\% & BLEU-4 & SER\% & BLEU-4 \\
              \hline
              \finetune & 49.94 &	0.382 &	44.00  & 0.375  \\
            \herding & 13.96 &	0.542 &	8.50  & 0.545  \\
            \random & 8.58 &	0.626 &	5.53 & 0.618   \\
           \loss & 8.21 &	0.684 &	5.20 & 0.669   \\
           \loss+\Ltwo  & 6.87 &	0.693 &	4.92  & 0.661  \\
           \loss+\distillation & 10.59 &	0.664 &	10.87  & 0.649   \\
            \loss+\dropout & 6.32 &	0.689 &	5.55  & 0.658   \\
            \arer & \textbf{3.63} &	\textbf{0.701} &	\textbf{3.52}  & \textbf{0.685}  \\
            \hline
                    \full & 3.08 &	0.694 &	2.98  & 0.672  \\
			\hline
		\end{tabular}
		\vspace{-0.05in} \caption{Performance of continually learning 7 DA intents using 250 exemplars. Best Performance excluding ``\full'' are in bold.}
        \label{table:da} \vspace{-0.05in}
	\end{table}

%% file: table/ablation.tex
    \begin{table}[t!]
   \fontsize{10}{12}\selectfont
    \setlength{\tabcolsep}{5pt}
		\centering
		\begin{tabular}{l cc cc }
			\hline
            & \multicolumn{2}{c}{$\Omega_{all}$}  & \multicolumn{2}{c}{$\Omega_{first}$} \\
            \cmidrule(r){2-3} \cmidrule(r){4-5} 
            & SER\% & BLEU-4 & SER\% & BLEU-4  \\
              \hline
 \arer & 4.82 & 0.592 & 3.88 & 0.569  \\
 \hline
 w/o ER & 6.41 &	0.584 &	5.85 &	0.559   \\
 w/o PE & 5.53	& 0.587 &	5.85 &	0.562  \\
 w/o AR & 5.57 & 0.587 & 	4.57 &	0.563 \\
%    \emph{AR} & 6.41 &	0.584 &	5.85 &	0.559   \\
%    \emph{ARER} & 5.53	& 0.587 &	5.85 &	0.572  \\
%    \emph{RPER} & 5.57 & 0.587 & 	4.57 &	0.573 \\
  %  \arer & 5.24 &	0.590	& 3.70 &	0.576  \\
			\hline
		\end{tabular}
		\vspace{-0.05in} \caption{Ablation study for \arer. ER / PE / AR stands for the Exemplar Replay loss / Prioritized Exemplars / Adaptive Regularization, respectively.} \vspace{-0.1in}
        \label{table:ablation}
	\end{table}

%% file: table/generation_examples.tex
\newcommand*{\red}{\textcolor{red}}
\newcommand*{\blue}{\textcolor{blue}}
\newcommand*{\purple}{\textcolor{purple}}
\newcommand*{\orange}{\textcolor{orange}}

	\begin{table*}[t!]
		\centering
		%\fontsize{10}{13}\selectfont
		\tabcolsep=0.2cm
		\begin{tabular}{c|p{12.5cm}}
			\hline
			\multicolumn{2}{c}{{Recommend (\textbf{Addr}=\textbf{regent street}, \textbf{Fee}=\textbf{free}, \textbf{Name}=\textbf{Downing College})}} \\ \hline
			Reference & \textit{[\textbf{Downing College}] is my favorite. It's located on [\textbf{regent street}] and it's [\textbf{free}] to get in}.  \\ \hline
		
			\multirow{ 2}{*}{\loss+\dropout} & \textit{[\textbf{Downing College}] is located in the city and it's located in the [\textbf{regent street}]. \purple{it's located at located at! it's located} in the \orange{[\textbf{Slot-Hotel-Area}]}.} \blue{(\textbf{missing: Fee=fre})} \\\hline
			\multirow{ 2}{*}{\arer} & \textit{I would recommend [\textbf{Downing College}]. It is located at [\textbf{regent street}] and has a entrance fee of [\textbf{free}].}  \red{(\textbf{correct})}\\ \hline
            
            \multicolumn{2}{c}{{ Recommend (\textbf{Area}=\textbf{centre of town}, \textbf{Name}=\textbf{saints church}, \textbf{Type}=\textbf{architecture destination})}} \\ \hline
			\multirow{ 2}{*}{Reference} & \textit{There is a [\textbf{saints church}] that is an [\textbf{architecture destination}] in the [\textbf{centre of town}], would you like that?} \\ \hline
			\multirow{ 2}{*}{\loss+\dropout} & \textit{I recommend [\textbf{saints church}] in the [\textbf{centre of town}]. \purple{it is a nice.  it is a guest house in a in a} \orange{[\textbf{Slot-Restaurant-Food}]}. }\blue{(\textbf{missing: Type=architecture destination})}\\\hline
			
			\arer & \textit{[\textbf{saints church}] is a [\textbf{architecture destination}] in the [\textbf{centre of town}].} \red{(\textbf{correct})}\\ \hline

		\end{tabular}
		\caption{Sample utterances generated for the first domain (``\textit{Attraction}'') after the NLG is continually trained on all 6 domains using 250 exemplars. Redundant and missing slots are colored in \orange{orange} and \blue{blue} respectively. Obvious grammar mistakes (redundant repetitions) are colored in \purple{purple}.}\vspace{-0.05in}
		\label{table:samples}
	\end{table*}

%% file: table/scvae.tex
% \begin{table}[t!]
% \fontsize{10}{12}\selectfont
%  \setlength{\tabcolsep}{3pt}
% 		\centering
% 		\begin{tabular}{l cc cc }
% 			\hline
%            % &  \multicolumn{4}{c}{\textbf{250 exemplars in total}}  \\
%             & \multicolumn{2}{c}{$\Omega_{all}$}  & \multicolumn{2}{c}{$\Omega_{first}$}  \\
%             \cmidrule(r){2-3} \cmidrule(r){4-5} 
%             & SER\% & BLEU-4 & SER\% & BLEU-4 \\
%               \hline
%               \finetune  & 60.83 &	0.361 &	98.86 &	0.186 \\
% 	%\surprise & - &	- &	-  & -   & - &	- &	-  & - \\
%     \herding  & 17.95 &	0.516 &	11.48  & 0.480  \\
%     \random & 9.31 &	0.543 &	7.52 & 0.483 	 \\
%    \loss  & 8.92 &	0.588 &	6.16  & 0.542  \\
%    \loss+\Ltwo   & 12.47 &	0.584 &	6.67  & 0.549  \\
%    \loss+\distillation  & 6.32 &	0.588 &	6.09  & 0.567  \\
%     \loss+\dropout  & 8.01 &	0.592 &	8.77  & 0.541 	 \\
%     \arer & \textbf{4.45} &	\textbf{0.595} &\textbf{4.04}  & \textbf{0.577} 	 \\
%     	 \hline
%             \full & 3.99 &	0.601 &	4.03  & 0.571  \\
% 			\hline
% 		\end{tabular}
% 		\vspace{-0.05in} \caption{Performance of using SCVAE as $f_{\theta}$. Best Performance excluding ``\full'' are in bold.} \vspace{-0.05in}
%         \label{table:scvae}
% 	\end{table}
    
    \begin{table}[t!]
\fontsize{10}{12}\selectfont
 \setlength{\tabcolsep}{4.5pt}
		\centering
		\begin{tabular}{l cc cc }
			\hline
           % &  \multicolumn{4}{c}{\textbf{250 exemplars in total}}  \\
            & \multicolumn{2}{c}{SCVAE}  & \multicolumn{2}{c}{GPT-2}  \\
            \cmidrule(r){2-3} \cmidrule(r){4-5} 
            & $\Omega_{all}$  & $\Omega_{first}$ & $\Omega_{all}$ & $\Omega_{first}$ \\
              \hline
              \finetune  & 60.83 &	98.86 & 28.69 &	31.76  \\
    \herding  & 17.95 &	11.48 & 11.95 &	10.48  \\
    \random & 9.31 & 7.52 & 9.87 & 8.85 \\
   \loss  & 8.92 &	6.16 & 8.72 &	8.20  \\
   \loss+\Ltwo   & 12.47  &	6.67  & 10.51  & 9.20  \\
   \loss+\distillation  & 6.32 & 6.09   & 8.41 & 8.09   \\
    \loss+\dropout  & 8.01 & 8.77 & 7.60 & 7.72  \\
    \arer & \textbf{4.45} & \textbf{4.04} & \textbf{5.32} & \textbf{5.05}  \\
    	 \hline
            \full & 3.99 & 4.03 & 4.75 & 4.53   \\
			\hline
		\end{tabular}
		\vspace{-0.05in} \caption{SER in $\%$ of using SCVAE and GPT-2 as $f_{\theta}$. Best Performance excluding ``\full'' are in bold.} \vspace{-0.08in}
        \label{table:scvae}
	\end{table}

%% file: text/conclusion.tex
\section{Conclusion}

In this paper, we study the practical continual learning setting of language generation in task-oriented dialog systems.
To alleviate catastrophic forgetting, we present \arer~which replays representative and diverse exemplars selected in a prioritized manner, and employs an adaptive regularization term based on EWC (Elastic Weight Consolidation). Extensive experiments on MultiWoZ-2.0 in different continual learning scenarios reveal the superior performance of \arer~.
The realistic continual learning setting and the proposed technique may inspire further studies towards building more scalable task-oriented dialog systems.

%% file: text/appendix.tex
\appendix

\part*{ Appendix}

\section{Reproducibility Checklist}

\subsection{Model Details and Hyper-parameters}
\label{appendix:model}

We first elaborate implementation details of the knowledge distillation (KD) baseline compared in our paper. We used the below loss term:
\begin{equation*}
\begin{array}{c}
    L_{KD}(\mathbf{Y}, \mathbf{d} , f_{\theta_{t-1}}, f_{\theta_t}) =  - \sum\limits_{k=1}^{K} \sum\limits_{i=1}^{{|L|}} \hat{p}_{k,i} \cdot log(p_{k,i} )
%     \hat{\pi}_i = \frac{ e^{\hat{o}_i} / T }{ \sum_{j=1}^N  e^{\hat{o}_j} / T}, \quad {\pi}_i = \frac{ e^{{o}_i} / T }{ \sum_{j=1}^N  e^{{o}_j} / T}, 
    \end{array}
\end{equation*}
where $L$ is the vocabulary that appears in previous tasks but not in task $t$.
At each position $k$ of $\mathbf{Y}$,  $ [\hat{p}_{k,1}, \dots, \hat{p}_{k,|L|}]$ is the predicted distribution\footnote{The temperature in \cite{hinton2015distilling} is set to 1 due to its minimum impact based on our experiments.} over $L$ given by $f_{\theta_{t-1}}$, and $[{p}_{k,1}, \dots, {p}_{k,|L|}]$  is the distribution given by $f_{\theta_t}$. 
$L_{KD}$ penalizes prediction changes on the vocabularies specific to earlier tasks. 
For all $\{\mathbf{Y}, \mathbf{d} \} \in \mathbf{D}_t \cup \mathbf{E}_{1:t-1}$, $L_{KD}$ is linearly interpolated with $L_{ER}$ by $L_{ER} + \eta \cdot L_{KD}$, with the $\eta$ tuned as a hyper-parameter . 

Hyper-parameters of SCVAE reported in Section \ref{subsec:scave} are set by default according to \url{https://github.com/andy194673/nlg-scvae}, except that the learning rate is set to 2e-3. 
For GPT-2, we used the implementation pipeline from \url{https://github.com/pengbaolin/SC-GPT}. We pre-processed the dialog act $\mathbf{d}$ into the format of : $\mathbf{d'} = [\; \mathbf{I} \;(\; s_1 = v_1, \dots, s_p = v_p\;)\; ]$, and the corresponding utterance $\mathbf{Y}$ is appended to be $\mathbf{Y'}$ with a special start token [BOS] and an end token [EOS]. $\mathbf{d'}$ and $\mathbf{Y'}$ are concatenated before feeding into GPT-2. The learning rate of Adam optimizer is set to 5e-5 without weight decay. As GPT-2 converges faster, we train maximum 10 epochs for each task with early stop applied to 3 consecutive epochs.

%For methods using exemplars, both training and validation set are continually expanded with exemplars extracted from previous tasks. 
Hyper-parameters of different methods are tuned to maximize SER$_{all}$ using grid search, and the optimal settings of different methods in various experiments are summarized in Table \ref{table:hyper}. 
%Hyper-parameters of SCVAE are set by default\footnote{\url{https://github.com/andy194673/nlg-scvae}}, except that the learning rate is set to 2e-3. 

\subsection{Domain Order Permutations}
\label{appendix:order}

\input{table/hyper}

\input{table/task_order}

In Table \ref{table:task_order}, we provide the exact domain order permutations of the 6 runs used in the experiments in Table \ref{table:domain_overall} and Figure \ref{fig:domain}.

\subsection{Computation Resource}

\input{table/computation}

All experiments are conducted using a single GPU (GeForce GTX TITAN X).
In Table \ref{table:computation}, we compared the average training time of one epoch using different methods.
\full\ spends more than 200s of extra computation overhead per epoch than other methods using bounded exemplars. 
\arer\ takes a slightly longer time to train than the methods except for \full.
Nevertheless, considering its superior performance, we contend that \arer\ achieves desirable resource-performance trade-off.
In addition, 250 exemplars are less than 1\% of historical data at the last task, and the memory usage to store a small number of exemplars is trivial.

\begin{figure*}[t!]
        \centering
        \includegraphics[width=0.92\textwidth]{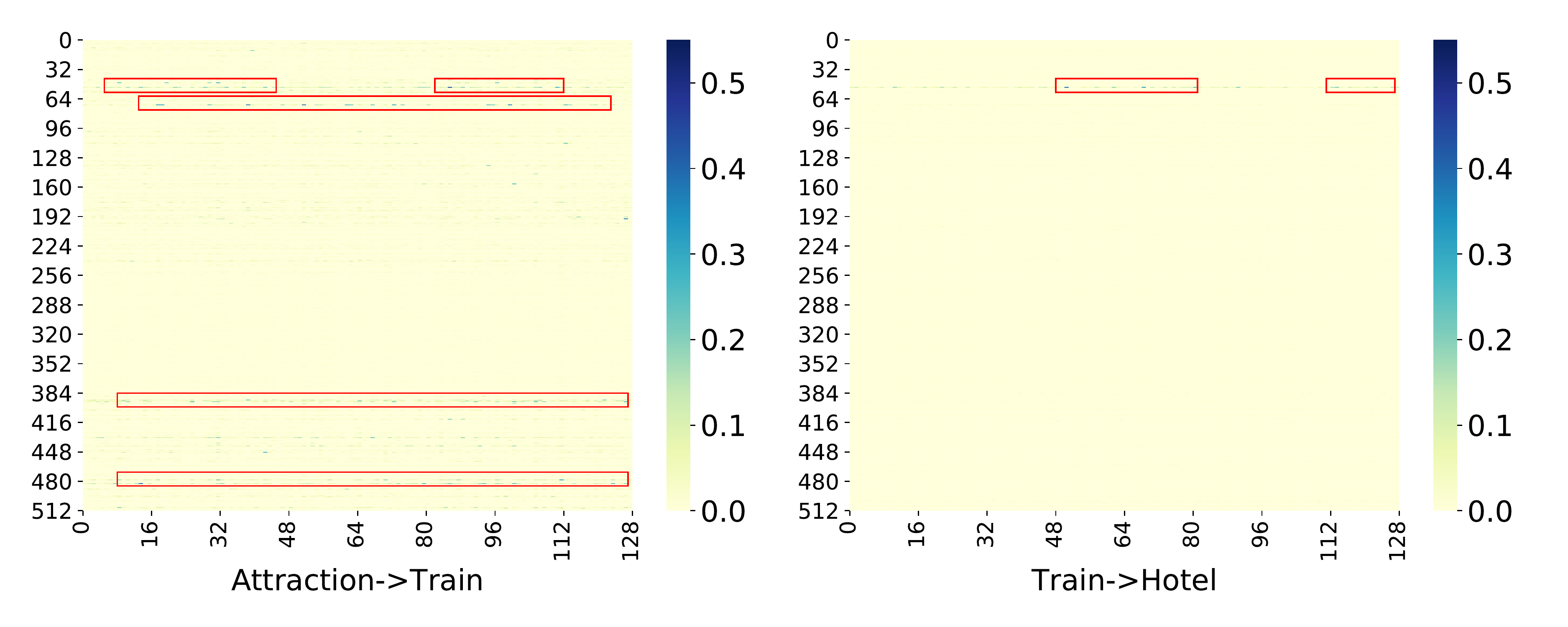}  \\
        \includegraphics[width=0.92\textwidth]{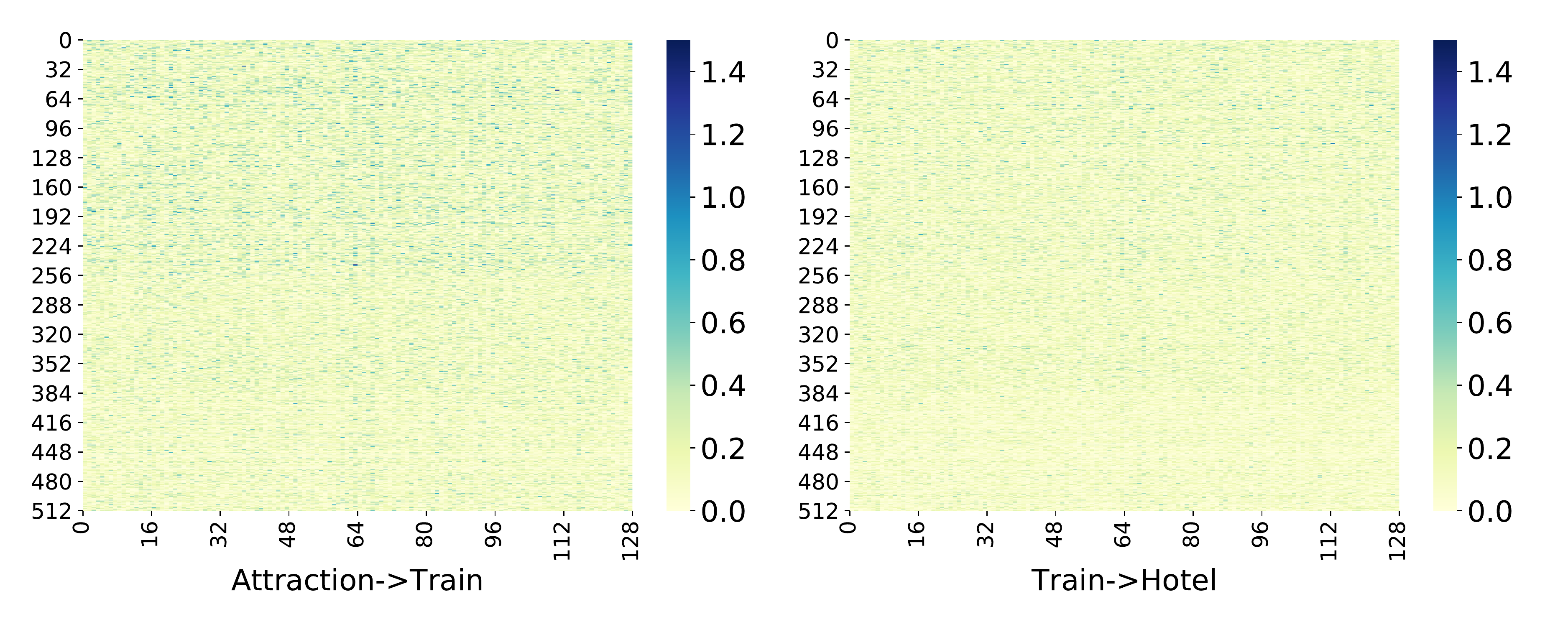}
\vspace{-0.05in} \caption{An visualization of the change of SCLSTM's hidden layer weights obtained from two consecutive tasks of \arer~(\textbf{Top}) and  \loss+\dropout~(\textbf{Bottom}). Two sample task transitions (``from \textit{Attraction}'' to ``\textit{Train}'', and then from  ``\textit{Train}'' to ``\textit{Hotel}'') are shown. High temperature areas of \arer\ is highlighted by red bounding boxes for better visualization.} \vspace{-0.05in}
        \label{fig:xxx}
   \end{figure*}    

\section{Supplementary Empirical Results}

% \subsection{Case Study}

% Table \ref{table:samples} shows two examples generated by \arer\ and the closest competitor (\loss+\dropout) on the first domain (“\textit{Attraction}”) after the NLG model is continually trained on all 6 domains starting with ``\textit{Attraction}''.
% In both examples, \loss+\dropout\ fails to generate slot ``Fee'' or ``Type'', instead, it mistakenly generates slots belonging to later domains (``\textit{Hotel}'' or ``\textit{Restaurant}'') with several obvious redundant repetitions colored in purple. It means that the NLG model is \textit{interfered} by utterance patterns in later domains, and it forgets some old patterns it has learned before.
% In contrast, \arer\ succeeds in both cases without forgetting previously learned patterns.

\subsection{Comparison to Pseudo Exemplar Replay}

\input{table/per}

%Another group of exemplar replay approach is ``pseudo'' exemplar replay. 
Instead of storing raw samples as exemplars, \citet{shin2017continual,riemer2019scalable} generate ``pseudo‘’ samples akin to past data.
The NLG model itself can generate pseudo exemplars. In this experiment, we replace the 500 raw exemplars of \random, \loss, and \arer\ by pseudo samples generated by the continually trained NLG model using the dialog acts of the same raw exemplars as input. Result comparing using pseudo or raw exemplars to continually learn 6 domains starting with ``\textit{Attraction}'' are illustrated in Table \ref{table:per}.
We can see that using pseudo exemplars performs better for \random, but worse for \loss\ and \arer. It means that pseudo exemplars are better when exemplars are chosen randomly, while carefully chosen exemplars (c.f. algorithm 1) is better than pseudo exemplars.
Explorations on utilizing pseudo exemplars for NLG is orthogonal to our work, and it is left as future work.

\subsection{Flow of Parameters Update}

To further understand the superior performance of \arer, we investigated the update of parameters throughout the continual learning process.
Specifically, we compared SCLSTM's hidden layer weights obtained from consecutive tasks, and the pairwise $L_1$ difference of two sample transitions is shown in Figure \ref{fig:xxx}. 

We can observe that \loss+\dropout\ tends to update almost all parameters, while \arer\ only updates a small fraction of them.
Furthermore, \arer\ has different sets of important parameters for distinct tasks, indicated by different high-temperature areas in distinct weight updating heat maps.
In comparison, parameters of \loss+\dropout\ seem to be updated uniformly in different task transitions.
The above observations verify that \arer\ indeed elastically allocates different network parameters to distinct NLG tasks to mitigate catastrophic forgetting.

%% file: table/hyper.tex
 \begin{table}[t!]
 \fontsize{10}{13}\selectfont
  \setlength{\tabcolsep}{1.5pt}
		\centering
		\begin{tabular}{l c c  }
			\hline
           % &  \multicolumn{4}{c}{\textbf{250 exemplars in total}}  \\
            & \textbf{Domains} & \textbf{DA Intents} \\
              \hline
   \loss~($\beta$) & 0.5/0.5/0.5/0.5 &	0.5 \\
   \Ltwo~(weight on L2) & 1e-3/1e-3/1e-3/5e-4 &	1e-2  \\
   \distillation~(weight on $L_{KD}$) & 2.0/3.0/2.0/0.5 & 5.0  \\
    \dropout~(rate) & 0.25/0.25/0.25/0.1 &	0.25 \\
    \arer~($\lambda_{base}$) & 300k/350k/200k/30k &	100k \\
			\hline
		\end{tabular}
		\caption{Optimal hyper-parameters of methods experimented in this paper. Four different values in the column ``Domains'' correspond to using 250 exemplars in both Table \ref{table:domain_overall} and Table \ref{table:da} / 500 exemplars in Table \ref{table:domain_overall} / using SCVAE / GPT-2 as $f(\theta)$ in Table \ref{table:scvae}, respectively. }
        \label{table:hyper} \vspace{-0.02in}
	\end{table}

%% file: table/task_order.tex
% \begin{table}[hbt!]
% \fontsize{10}{13}\selectfont
% \begin{tabular}{|l|l|l|l|l|l|l|}
% \hline
% Attraction & 0 & 2 & 1 & 3 & 4 & 3 \\
% Booking    & 4 & 0 & 3 & 4 & 2 & 4 \\
% Hotel      & 3 & 5 & 0 & 1 & 1 & 2 \\
% Restaurant & 4 & 4 & 2 & 0 & 5 & 1 \\
% Taxi       & 5 & 1 & 4 & 2 & 0 & 5 \\
% Train      & 1 & 3 & 5 & 5 & 3 & 0 \\
% \hline
% \end{tabular}
% \caption{Task orders of 6 experiments. Each column illustrates an experiment and each entry represents the relative order of the task in corresponding experiment.}
% \label{table:task_order}
% \end{table}

\begin{table}[t!]
\centering
\fontsize{10}{13}\selectfont
\begin{tabular}{c | cccccc}
\hline
Run 1  & 0 & 5 & 2 & 1 & 3 & 4 \\
\hline
Run 2    & 1 & 4 & 0 & 5 & 3 & 2 \\
\hline
Run 3     & 2 & 0 & 3 & 1 & 4 & 5 \\
\hline
Run 4     & 3 &2 &4 &0 & 1 & 5 \\
\hline
Run 5       & 4 & 2 & 1 & 5 & 0 & 3 \\
\hline
Run 6      & 5 & 3 & 2 & 0 & 1   &4 \\
\hline
\end{tabular}
\caption{Each row corresponds to a domain order permutation. The mapping from domain to id is: \{``\textit{Attraction}'': 0,  ``\textit{Booking}’‘: 1,  ``\textit{Hotel}'': 2, ``\textit{Restaurant}'': 3, ``\textit{Taxi}'': 4, ``\textit{Train}'': 5.\}}
\label{table:task_order}
\end{table}

%% file: table/computation.tex
%  \begin{table}[htb!]
%  \fontsize{10}{12}\selectfont
%   \setlength{\tabcolsep}{3pt}
% 		\centering
% 		\begin{tabular}{l c c  }
% 			\hline
%            % &  \multicolumn{4}{c}{\textbf{250 exemplars in total}}  \\
%             & \textbf{Computation Time} & \textbf{Memory Usage} \\
%               \hline
%    \finetune & 18 &	- \\
%    \distillation & 32 & -  \\
%     \dropout & 22 &	- \\
%     \arer & 39 &	- \\
%     \full & 224 &	- \\
% 			\hline
% 		\end{tabular}
% 		\caption{Average computation time of one epoch and memory usage at the last continual learning task in Section \ref{subsec:domain} using250 exemplars}
%         \label{table:computation}
% 	\end{table}
    
     \begin{table}[htb!]
 \fontsize{10}{13}\selectfont
  \setlength{\tabcolsep}{2pt}
		\centering
		\begin{tabular}{c c c c  c c c}
              \hline
   $\finetune$ & \loss & $\Ltwo$ & $\distillation$ & $\dropout$ & $\arer$ & $\full$  \\
   \hline
   17.5s & 18.5s & 19.5s & 24.6s & 15.5s & 39.5s & 242.5s \\
			\hline
		\end{tabular}
		\caption{Average training time of one epoch at the last task when continually learning 6 domains starting with ``\textit{Attraction}'' using 250 exemplars. Methods other than \finetune\ and \full\ are applied on top of \loss.} \vspace{-0.1in}
        \label{table:computation}
	\end{table}

%% file: table/per.tex
\begin{table}[hbt!]
\fontsize{10}{13}\selectfont
 \setlength{\tabcolsep}{2pt}
		\centering
		\begin{tabular}{l cc cc }
			\hline
           % &  \multicolumn{4}{c}{\textbf{250 exemplars in total}}  \\
            & \multicolumn{2}{c}{$\Omega_{all}$}  & \multicolumn{2}{c}{$\Omega_{first}$}  \\
            \cmidrule(r){2-3} \cmidrule(r){4-5} 
            & SER\% & BLEU-4 & SER\% & BLEU-4 \\
              \hline
    \random  & 9.82 &	0.495 &	8.64 &	0.405 \\
   \emph{Pseudo}-\random  & 9.26 &	0.551 &	6.88 &	0.519\\
   \hline
   \loss  & 7.84 &	0.573 &	6.20 &	0.523  \\
    \emph{Pseudo}-\loss &  8.87 &	0.557 &	6.37 &	0.521 \\
    \hline
    \arer  & 4.43 &	0.597 &	3.40 &	0.574 	 \\
    \emph{Pseudo}-\arer & 5.07 &	0.590 &	3.51 &	0.570 	 \\
			\hline
		\end{tabular}
		\caption{Comparison with Pseudo Exemplar Replay. } \vspace{-0.1in}
        \label{table:per}
	\end{table}